\documentclass[10pt,twocolumn,letterpaper]{article}

\usepackage{caption}
\usepackage{subcaption}
\usepackage{wrapfig}
\usepackage{graphicx}

\usepackage{multirow}
\usepackage{booktabs}
\usepackage{colortbl}
\usepackage{longtable} 
\usepackage{array}

\usepackage{amsmath}
\usepackage{amssymb}
\usepackage{mathtools}
\usepackage{amsthm}
\usepackage{nicefrac} 
\usepackage{derivative} 
\usepackage{empheq} 

\usepackage{bm} 
\usepackage{mathrsfs} 
\usepackage{dsfont} 
\usepackage{pifont}
\usepackage{fontawesome5} 
\usepackage{latexsym} 
\usepackage{marvosym}  
\usepackage{lmodern}

\usepackage{lipsum}

\usepackage[dvipsnames]{xcolor}

\usepackage{comment}

\usepackage[english]{babel}
\usepackage{csquotes}

\usepackage{algorithm}
\usepackage{algorithmicx}
\usepackage[noend]{algpseudocode}
\makeatletter
\renewcommand{\ALG@beginalgorithmic}{\normalsize} 
\makeatother

\usepackage{tablefootnote}

\usepackage{tikz}
\usepackage{circuitikz}

 \usepackage{cvpr}              

\usepackage{algorithm}

\newcommand{\red}[1]{{\color{red}#1}}

\definecolor{cvprblue}{rgb}{0.21,0.49,0.74}
\usepackage[pagebackref,breaklinks,colorlinks,allcolors=cvprblue]{hyperref}

\def\confName{CVPR}
\def\confYear{2026}

\title{Towards Robust Content Watermarking Against Removal and Forgery Attacks}

\author{ 
Yifan Zhu \thanks{These authors contributed equally to this work.} \\
AMSS and UCAS, CAS \thanks{Academy of Mathematics and Systems Science, Chinese Academy of Sciences and University of Chinese Academy of Sciences.} \\
{\tt\small zhuyifan@amss.ac.cn}
\and
Yihan Wang \footnotemark[1] ~ \thanks{Corresponding authors.} \\
University of Waterloo \\
{\tt\small yihan.wang@uwaterloo.ca}
\and
Xiao-Shan Gao \footnotemark[3]\\
AMSS and UCAS, CAS \footnotemark[2]  \\
{\tt\small xgao@mmrc.iss.ac.cn}
}

\begin{document}
\maketitle

\begin{abstract}
 Generated contents have raised serious concerns about copyright protection, image provenance, and credit attribution. A potential solution for these problems is watermarking. Recently, content watermarking for text-to-image diffusion models has been studied extensively for its effective detection utility and robustness. However, these watermarking techniques are vulnerable to potential adversarial attacks, such as removal attacks and forgery attacks. In this paper, we build a novel watermarking paradigm called \textit{Instance-Specific watermarking with Two-Sided detection} (\textbf{ISTS}) to resist removal and forgery attacks. Specifically, we introduce a strategy that dynamically controls the injection time and watermarking patterns based on the semantics of users' prompts. Furthermore, we propose a new two-sided detection approach to enhance robustness in watermark detection. Experiments have demonstrated the superiority of our watermarking against removal and forgery attacks.
\end{abstract}

\section{Introduction}\label{sec:introduction}
Diffusion models have become the dominant class of generative models for visual tasks, powering many popular text-to-image and text-to-video systems such as Stable Diffusion~\citep{RombaBLEO21}, DALL-E~\citep{RamesPGGVRCS21}, Midjourney, and Sora~\citep{LiuZLYGCYHSGHS24}. Content produced by these models has become widespread across social media and even traditional publications~\citep{Liu2022}. Notably, a synthetically generated image has even won a world photography award~\citep{Glynn2023}.

The photorealistic quality of such generated media has raised serious concerns about intellectual property protection, including issues of image provenance, authorship, and credit attribution. Watermarking has emerged as a promising solution for verifying whether an image originates from a specific diffusion model~\citep{WenKGG23,YangZCFZY24,GunnZS24}. In this approach, identity markers are embedded into the generated outputs during the synthesis process without degrading their semantics or perceptual quality. During detection, model providers can confirm the provenance of an image by detecting or recovering these markers.
Researches on watermarking for diffusion models have grown rapidly in recent years, with many works focusing on improving detection accuracy, image fidelity, and robustness against natural variations such as transformations, noise, or compression~\citep{ZhangLiBBG24,CiYSS25,WangXWDLLWR25}.

However, since ownership and credit for generated content can carry reputational and even financial implications, malicious adversaries have strong incentives to circumvent watermarking mechanisms. Such attackers may attempt to remove identity markers from watermarked images or to forge watermarked images from unmarked ones. These attacks often exploit gradients propagated through the diffusion process and the statistical properties of large sets of watermarked samples, achieving high success rates in watermark removal and forgery~\citep{MulleLTFQ24,YangCSS24,JainKMTSMCMT25a,kassis2025unmarker}.
Such vulnerability presents severe challenges to the reliability of current watermarking methods.
To address these challenges, we propose \textit{Instance-Specific watermarking with Two-Sided detection} (\textbf{ISTS}), which achieves enhanced resilience against forgery and removal attacks while maintaining comparable utility in non-adversarial settings. 

%
Our key insight is that, despite existing attacks being frequently characterized by black-box access, in which the adversary does not know the underlying generative model or watermarking approach, the prevailing use of static, single-type watermarking methods effectively grants the attacker additional prior knowledge. 
For instance, the well-known Tree-Ring watermarking~\citep{WenKGG23} injects a fixed ring pattern into the Fourier space of latent noises for all prompts.
This unintended leakage allows adversaries to more easily infer the watermark's structural characteristics, facilitating its removal or the creation of forged watermarks.
To mitigate this vulnerability, we propose a strategy that customizes both the watermark pattern and the injection time for each generation.
Concretely, given a prompt, we first generate a non-watermarked image and encode it into a semantic feature vector using the CLIP encoder. 
We then determine the watermarking parameters, such as the pattern position and the injection time step, based on this semantic vector through a pretrained semantic-based selector.
Finally, using the selected parameters, we generate the corresponding watermarked image from the same prompt.
For detection, we first encode the suspicious image into its semantic feature vector, retrieve the corresponding watermarking parameters, and then compare the target region with the predefined watermark pattern.
Since watermark injection minimally affects image semantics, the semantic features of watermarked and non-watermarked images remain closely aligned, ensuring that the recovered parameters are consistent with those used during the watermark injection process.

In addition, we observe that the commonly adopted one-sided detection introduces a critical vulnerability to removal attacks. To address this issue, we propose a two-sided detection scheme that captures opposite latent representations, effectively strengthening robustness against such attacks while preserving the watermark's utility.

We conduct experiments on various content watermarking with our ISTS against three representative removal attacks and three forgery attacks. Evaluation results demonstrate the superior performance of our proposed ISTS, achieving state-of-the-art detection AUC and TPR@1\%FPR across removal and forgery attacks in both average and worst-case scenarios.

\begin{figure*}[t]
    \centering
    \includegraphics[width=0.9\textwidth]{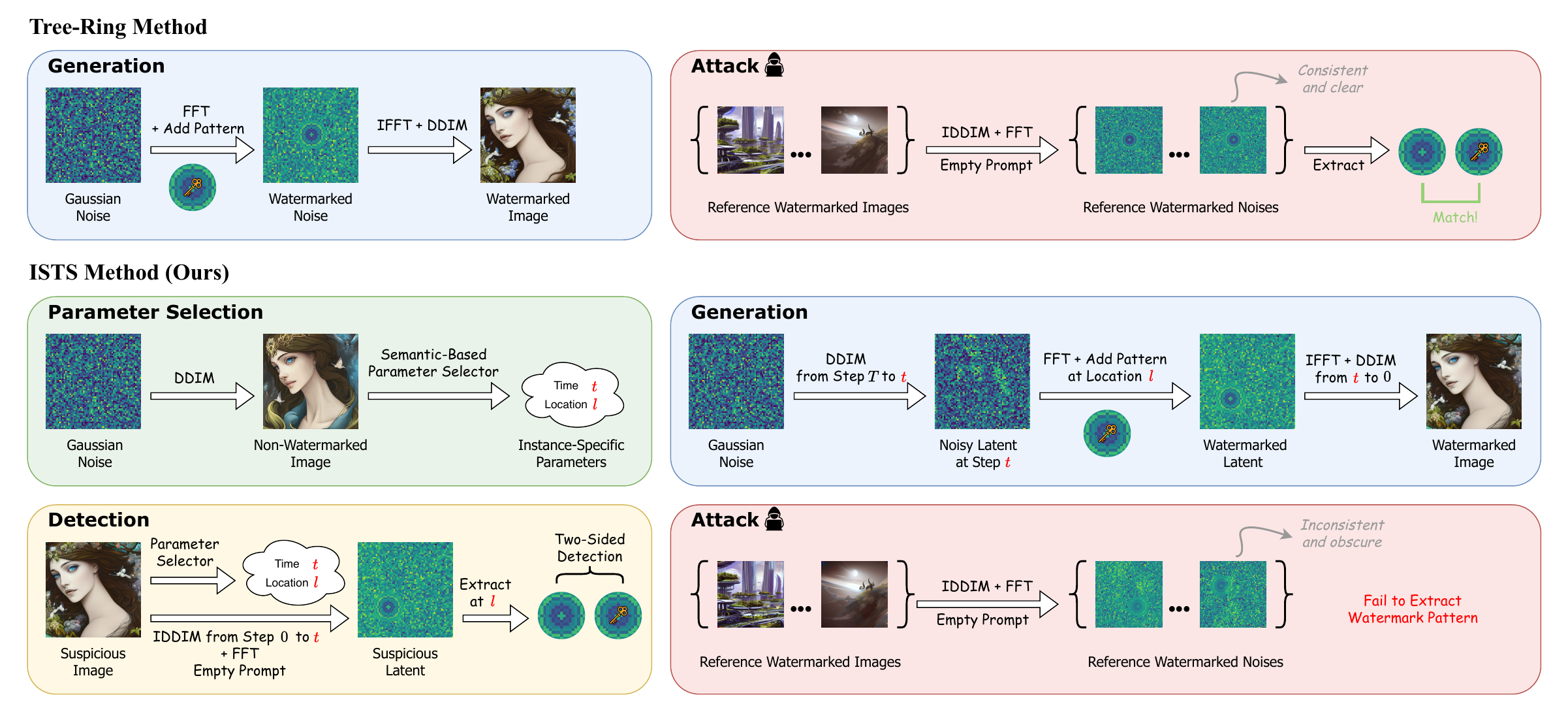}
    \caption{Overview of our ISTS method. The \textbf{top} row shows the original Tree-Ring watermarking. To generate watermarked images, it injects tree-ring patterns in the center of the frequency domain at the initial noisy space. Such a static scheme leaks information about the watermark pattern and thus exposes high vulnerability to removal and forgery attacks. 
    The \textbf{bottom} row shows our dynamic approach. For each generation, we first generate a non-watermarked image and decide time and location parameters $(t,l)$ using \textit{a semantic-based selector}.
    Then, we execute the first $T-t$ steps of DDIM, inject watermarks at coordinate $l$ in the frequency domain, and continue the rest of DDIM generation. During detection, we retrieve the parameters from a suspicious image and extract the watermarking area, followed by \textit{two-sided detection}.
    In adversarial settings, an attacker unaware of time and location parameters fails to extract the ground-truth watermark patterns from reference images. Thus, the proposed approach achieves enhanced robustness against removal and forgery attacks.
    }
    \label{fig:dyn_wm}
\end{figure*}

\section{Related Works} \label{sec:related_works}

\textbf{Watermarking.}
Watermarking on generative models has been widely studied in recent years. In text-to-image generative models, a representative watermarking method called Tree-Ring~\citep{WenKGG23} injects a specific ring pattern into the Fourier space of the latent representation, achieving higher robustness compared with traditional image watermarking methods~\citep{cox2002digital,zhang2019robust,tancik2020stegastamp}. Followed up with many content watermarking methods, such as ZoDiac~\citep{ZhangLiBBG24}, RingID~\citep{CiYSS25}, ROBIN~\citep{huang2024robin}, Gaussian-Shading~\citep{YangZCFZY24}, and Shallow Diffuser~\citep{li2024shallow}. More works~\citep{guo2024freqmark,yang2025gaussian, ArabiFWHC25,WangXWDLLWR25, pautov2025spread} have been proposed to further enhance content watermarking in diffusion models. 
Other works try to inject watermarking into text-to-image generative models by finetuning the model~\citep{zhao2023recipe,zhang2024sat,sun2025diffmark} or training a watermarking extractor~\citep{fernandez2023stable}. \citet{GunnZS24} tries to use pseudorandom error-correcting codes to inject an undetectable watermark into generative image models.
SEAL~\citep{ArabiWHC25} Using SimHash to control the randomness of the denoising process. 
In large language models, \citet{kirchenbauer2023watermark} first exploits the Red-Green list to imprint watermarks into the generated text. Further research on watermarking for language models has also been studied in recent years~\citep{kirchenbauer2023reliability, zhao2023provable, kuditipudi2023robust, christ2024undetectable,liu2024adaptive,xie2025debiasing}.
More broadly, watermarking in text-to-video generative models has been investigated recently~\citep{fernandez2024video,hu2025videoshield,hu2025videomark}; watermarking in generative tabular data and data poisoning attacks has also been considered~\citep{he2024watermarking,zheng2024tabularmark, zhu2025provable}.

\noindent \textbf{Removal and forgery attacks on watermarking.}
Removal attacks have been widely studied in the area of post-hoc image watermarking, including distortion methods~\citep{voloshynovskiy2001attacks,an2024benchmarking}, reconstruction and generation processes~\citep{li2019towards,cheng2020learned,zhao2023provable,liu2024image}, and learning-based approaches~\citep{jiang2023evading,hu2024transfer,lukas2023leveraging}. Forgery attacks against traditional image watermarking have also been investigated~\citep{kutter2000watermark,voloshynovskiy2001attacks,saberi2023robustness,dong2025wmcopier}. \citet{ba2025robust} utilizes channel-aware feature extraction to remove and forge image watermarks.
Content watermarking shows strong robustness against image distortion and diffusion purification~\citep{yoon2021adversarial, nie2022diffusion}, however, a recent study has found that content watermarking is vulnerable to removal and forgery attacks, even under simple averaging~\citep{YangCSS24}. 
~\citet{MulleLTFQ24} utilizes a surrogate diffusion model to remove content watermarking by optimizing the watermarked latent in the opposite direction and to forge content watermarking by approximating the watermark pattern in the latent space when a single watermarked image is leaked. \citet{JainKMTSMCMT25a} achieves removal and forgery attacks via the VAE encoder with a single watermarked image.
Recent works~\citep{jovanovic2024watermark,chen2024mark,zhang2024large} have demonstrated that watermarking in large language models is also fragile to removal (scrubbing) and forgery (spoofing) attacks.

In this paper, we mainly focus on content watermarking in text-to-image generative models, which have been well studied and are vulnerable to removal and forgery attacks~\citep{MulleLTFQ24,YangCSS24,JainKMTSMCMT25a}. We introduce a novel watermarking method, ISTS, to improve robustness against these attacks.

\section{Background}\label{sec:background}

\subsection{Preliminaries}
\textbf{Diffusion Models.}
Diffusion models~\citep{ho2020denoising, song2020score} have achieved remarkable success in the area of image generation. A representative framework called the Denoising Diffusion Probabilistic Model (DDPM)~\citep{ho2020denoising} progressively adds Gaussian noise $\epsilon\sim\mathcal{N}(0,I)$ to the original image $x_0$ through the following forward process:
\begin{equation}
    x_t=\sqrt{\alpha_t}x_0+\sqrt{1-\alpha_t}\epsilon
\end{equation}
with a certain noise schedule $\{\alpha_t\}$. We approximate the reverse process by estimating the noise $\epsilon_t$ using the diffusion model. To enhance sample efficacy, \citet{song2020denoising} proposed Denoising Diffusion Implicit Models (DDIM) to predict the previous state $x_{t-1}$ by:
\begin{align}
    x_{t-1}=\sqrt{\alpha_{t-1}}\left(\frac{x_t-\sqrt{1-\alpha_t}\epsilon_\theta(x_t,t)}{\sqrt{\alpha_t}}\right) \nonumber \\ +\sqrt{1-\alpha_{t-1}}\epsilon_\theta(x_t,t), \label{eq:ddim}
\end{align}
where $\epsilon_\theta$ is parameterized and trained to predict $\epsilon_t$ at time $t$. A merit of DDIM is that one can sample $x_t$ from $x_{t-1}$ through the DDIM inversion process, similar to \cref{eq:ddim}.

\noindent \textbf{Text-to-Image (T2I) Diffusion Models.}
Diffusion models can be generalized to text-to-image~\citep{rombach2022high, esser2024scaling} by embedding image $x$ into the latent space with the encoder $\mathcal{E}$ to obtain latent $z_0=\mathcal{E}(x)$, then guiding the image generation process with a text prompt $p$. We define the denoising process to be $$z_{t_1}=\mathcal{M}_{t_2\to t_1}(p, z_{t_2}),$$
where $\mathcal{M}_{t_2\to t_1}$ is to denoise from step $t_2$ to $t_1$ with the latent diffusion model $\mathcal{M}$ when $t_1<t_2$. After obtaining the denoised latent $\tilde{z}_0$, decode the latent using the decoder $\mathcal{D}$ to obtain the generated image $\tilde{x}_0=\mathcal{D}(\tilde{z}_0)$. To simplify the notation, we omit the decoder $\mathcal{D}$ in this paper when no ambiguity exists. For the inversion process, similarly, we denote it as 
\begin{equation}
z_{t_2}=\mathcal{M}^{\textup{Inv}}_{t_1\to t_2}(p, z_{t_1}), t_1<t_2.
\end{equation}
The classifier-free guidance (CFG)~\citep{ho2022classifier} has also been employed for the sampling process.

\noindent \textbf{Content watermarking.}
A well-known content watermarking technique in text-to-image diffusion models, called Tree-Ring~\citep{WenKGG23}, injects a watermark pattern into the initial latent noise vector $z_T$. Following Tree-Ring, many content watermarking methods have been developed, including Gaussian-Shading~\citep{YangZCFZY24}, ROBIN~\citep{huang2024robin}, RingID~\citep{CiYSS25}, ZoDiac~\citep{ZhangLiBBG24}, Shallow Diffuse~\citep{li2024shallow}, etc. 
These methods modify the generation process of text-to-image diffusion models without finetuning, achieving high robustness compared to existing image watermarking methods and providing a plug-and-play approach to inject effective watermarking. 

\noindent \textbf{Watermarking detector.} The watermark detector $D$ is a binary classifier, where $D(x)=1$ if the image $x$ is recognized as a watermarked instance and $D(x)=0$ if  $x$ is recognized as a non-watermarked instance.

\subsection{Threat Model of Removal and Forgery Attacks}

\noindent \textbf{Adversary's objectives.}
In our scenario, the adversary has two objectives, removal attacks and forgery attacks.

\noindent\textit{Removal Attacks}: In this setting, the adversary starts with a watermarked image generated by a T2I model. The objective is to slightly modify the image so that the resulting output bypasses watermark detection while preserving perceptual similarity. Formally, for a watermarked image $I^{w}$, removal attacks try to find the perturbation $\delta$, such that
\begin{align*}
    D(I^{w}+\delta)=0,\ \  d(I^{w}+\delta, I^{w})\leq\epsilon,
\end{align*}
where $d$ is a metric measuring the difference of two images.


\noindent \textit{Forgery Attacks}: The adversary begins with a benign (non-watermarked) image and seeks to alter it such that it is falsely recognized as watermarked during detection while preserving perceptual similarity.
Formally, given a benign image $I^{b}$, the forgery attacker aims to search for a perturbation $\delta$, such that
\begin{align*}
    D(I^b+\delta)=1,\ \  d(I^b+\delta, I^b)\leq\epsilon.
\end{align*}

\noindent \textbf{Adversary's knowledge and capability.}
The adversary is assumed to have no access to the weights of the target diffusion model, and is unaware of the watermarking algorithm or its implementation details. However, they can leverage
surrogate models to compute gradients, granting a strong capability that enables a range of gradient-based attacks \citep{MulleLTFQ24}, and can possess a reference image containing a watermark, which could be used in attempts to forge or replicate the watermark.
Furthermore, they may obtain some watermarked images to extract population characteristics~\citep{YangCSS24}.


\subsection{Failure of Existing Content Watermarking Against Removal and Forgery Attacks}

\begin{figure}[h]
    \centering
    \includegraphics[width=0.99\linewidth]{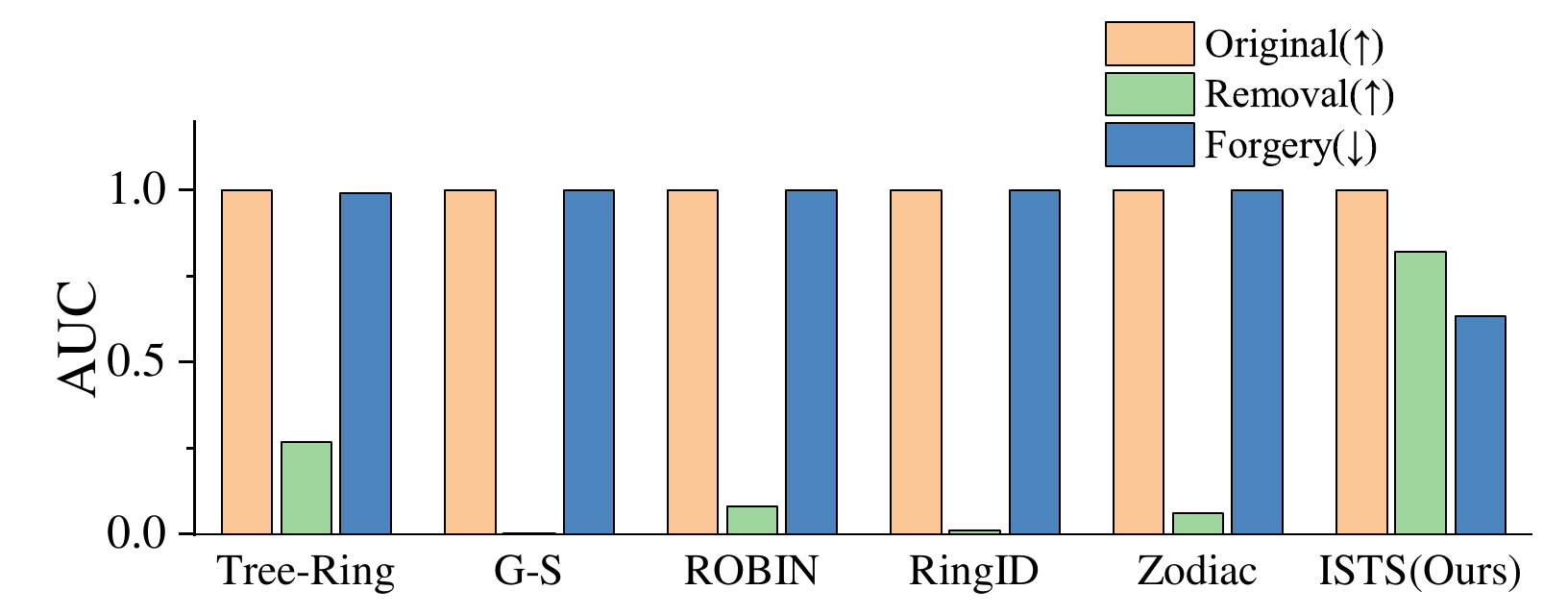}
    \caption{The detection AUC of benign watermarking, and watermarking after removal and forgery attacks~\citep{MulleLTFQ24}. 
    Results demonstrate that existing methods are vulnerable to removal attacks (lower AUC) and forgery attacks (higher AUC), while our ISTS is robust against them.
    }
    \label{fig:fail}
\end{figure}

\noindent Recent works~\citep{YangCSS24, MulleLTFQ24, JainKMTSMCMT25a} have shown that content watermarking is susceptible to removal and forgery attacks, even if only one watermarked image is leaked, pointing to significant risks associated with the real-world deployment of content watermarking. For example, as we showed in \Cref{fig:fail}, existing content watermarking can be almost completely destroyed by the removal and forgery attacks proposed by \citet{MulleLTFQ24}.  After the removal attack, the detector can achieve only less than 0.1 AUC between benign generated images and watermarked images under removal for Gaussian-Shading, ROBIN, RingID, and Zodiac watermarking, rendering them ineffective. 
After the forgery attack involving only one leaked watermarked image, benignly generated images can be modified into watermarked images, and the watermarking detector will confidentially view these forged images as watermarked images, with an AUC of nearly 1.0.
The failure of existing content watermarking facilitates our design of a new watermarking paradigm that is robust against such forgery and removal attacks.

\section{Methodology}\label{sec:method}

To overcome the vulnerability of watermarking against removal and forgery attacks, we propose a new watermarking method called \textit{Instant-Specific watermarking with Two-Sided detection} (\textbf{ISTS}), which
customizes both the watermark location and the injection time for each instance during generation, and conducts a two-sided test for detection. 

\subsection{Instance-Specific Watermarking}
\label{sec:dyn_pattern}
Existing content watermarking methods, such as Tree-Ring, RingID, and Zodiac, rely on a static watermarking pattern to assist in detection. The pattern is injected into the initial noisy latent space through a Fourier transformation, making it invisible to humans and gaining decent robustness under image distortions. 
However, such static, single-type watermarking grants potential attackers additional prior knowledge about the scheme. This unintended leakage enables them to infer the watermark’s structural characteristics, facilitating removal and forgery attacks. 
For instance, an adversary can reverse images to the latent space using existing methods, such as DDIM-Inversion~\citep{dhariwal2021diffusion}. 
Therefore, with a watermarked image for reference, the adversary can forge more watermarked images from benign ones by minimizing the distance between the reference and benign latent vectors with gradient descent~\citep{MulleLTFQ24}. 

\noindent \textbf{Dynamic patterns and injections.}
To enhance resilience to removal and forgery attacks, we introduce instance-specific parameters to control the watermark injected for each image generation, thereby increasing the variety of watermarks and limiting the efficacy of the adversary's operation.
For each image generation, we dynamically select parameters in both time and location dimensions, i.e., the injection step $t$ and the pattern coordinate $l$,  making it substantially harder for an adversary to predict or exploit them.

Specifically, as shown in \Cref{fig:dyn_wm}, we first generate a non-watermarked image, extract its features using a CLIP encoder, and map those features to time and location parameters $(t, l)$ through a semantic-based selector. Then, the watermark pattern is injected at coordinate $l$ in step $t$, as shown in \Cref{alg:our_wm} and Appendix \ref{app:exp_detail}. 
The parameter selector is obtained through K-Means clustering applied to non-watermarked images, as demonstrated in \Cref{alg:cls}, in which the parameter mapping $\phi$ is predefined based on the modulo operation, which will be demonstrated in Appendix \ref{app:exp_detail}.
Since the semantics of non-watermarked and watermarked images are similar, the paired images tend to be predicted to the same parameters, resulting in consistency for generation and detection, thereby ensuring strong detection performance even under adversarial conditions. 

Dynamic patterns and injections significantly enhance the correlation between image concepts and watermarking styles, further degrading the existing removal and forgery attacks that use a single watermarked image~\citep{MulleLTFQ24, JainKMTSMCMT25a} or the averaging of watermarked images~\citep{YangCSS24} as features to extract the watermarking pattern for each generated image.

\noindent \textbf{Resilience improvement.}
Recall that the adversary lacks knowledge of the watermarking algorithm's internal design and must perform a single-shot attack under an assumed standard setting.
Thus, the greater the discrepancies between the assumed and the actual configurations, the lower the attack's success rate. 
To illustrate how our proposed scheme benefits the resilience to removal and forgery attacks, we will analyze six concrete attacks.

Regarding forgery attacks, \citet{MulleLTFQ24} and \citet{JainKMTSMCMT25a} both consider the scenario in which an attacker has access to a single watermarked image for reference. 
Our approach requires the watermark to be placed according to the outputs of a semantic-based selector.
However, in these two attacks, the watermark pattern forged from the reference image does not match the semantics of the benign image, thereby resulting in a detection failure.
Moreover, they attempt to align the reference feature and the forged feature in the latent space.
However, the dynamic time parameter hinders the attack from tracing back to the exact injection step at which gradient-based optimization is conducted.
\citet{YangCSS24} considers a forgery attacker who obtains a multitude of non-watermarked and watermarked images.
The attacker extracts watermarking features by calculating the residuals of the averaged watermarked images and the averaged non-watermarked images.
Our proposed scheme diversifies the watermarking features residing in the pixel space.
Thus, the features in one watermarked image might offset those in other watermarked images, and their average fails to present meaningful information about any watermarking pattern.

Regarding removal attacks, \citet{JainKMTSMCMT25a} uses a plain image with all values equal to the mean of the targeted watermarked image as the object of optimization. However, this results in biased removal features, since our dynamic patterns generate distinct watermark features across different semantics, thereby substantially degrading their attack performance.
In addition, the gradient-based removal attack by \citet{MulleLTFQ24} suffers from an injection-step mismatch issue, similar to its forgery counterpart.
Furthermore, in \citet{YangCSS24}, the residuals between averaged non-watermarked and watermarked images no longer capture watermark features under our dynamic pattern. Consequently, their strategy of removing watermarks by subtracting the average fails to remain effective.


\begin{algorithm}[h]
\caption{Parameter Selector Training}
\label{alg:cls}
\begin{algorithmic}
\State{\bfseries Input:} 
Model $\mathcal{M}$, prompt set $\mathcal{P}$, CLIP extractor $g$, clustering number $N$, a parameter mapping $\phi$.
\State {\bfseries Output:} 
Parameter selector $f$.
\For{$p$ in $\mathcal{P}$}
\State Sample noise $z_T$ from $\mathcal{N}(0,I)$.
\State $I_p^c\gets \mathcal{M}(p, z_T)$. \Comment{Obtain non-watermarked images}
\State $z_p^c\gets g(I_p^c)$. \Comment{Extract features}
\EndFor
\State $y_p^c\gets \textup{K-Means}(z_p^c,\left\{z_p^c\right\}_{p\in\mathcal{P}},N)$. \Comment{Assign labels}
\State Train a classifier $h$ using labeled dataset $\left\{(z_p^c, y_p^c)\right\}_{p\in\mathcal{P}}$.
\Return $f\gets \phi\circ h\circ g$.
\end{algorithmic}
\end{algorithm}

\subsection{Two-Sided Detection}
Beyond the generation process, we find that the detection process could also expose the vulnerabilities of watermarking. 
Existing methods like Tree-Ring variants, typically rely on one-sided detection with the detection metric
\begin{align*}
    d=\frac{1}{|M|}\sum\limits_{i\in M}|W_i-\mathcal{F}(z_T)_i|,
\end{align*}
in that $W_i$ is the watermarking pattern under dimension index $i$ and $z_T=\mathcal{M}^{\textup{Inv}}_{0\to T}(z_0)$ for the suspect image latent $z_0=\mathcal{E}(x_0)$ under DDIM Inversion process~\citep{dhariwal2021diffusion} as proposed in Tree-Ring~\citep{WenKGG23}, $\mathcal{F}$ is the Fourier transformation.
This detection approach introduces a critical vulnerability to removal attacks. Specifically, an adversary can eliminate the watermark by optimizing the generated images to induce opposite latent representations (Details in Appendix \ref{app:removal_forgery}).
A simple yet effective removal attack proposed by \citet{MulleLTFQ24} has shown significant degradation of watermarking, as outlined in Table 1. It induces the watermarking latents into the opposite to generate the removed images. To overcome this issue, we modify our detection approach from one-sided to two-sided. With this simple yet effective change, our watermarking paradigm demonstrates stronger robustness against removal attacks while preserving detection efficacy in non-adversarial conditions.  Formally, the detection metric works as follows:
\begin{align*}
    d
    = \!\min \!\left\{\!\frac{1}{|M|}\!\sum\limits_{i\in M}\!|W_i-\mathcal{F}(z_T)_i|,\frac{1}{|M|}\!\sum\limits_{i\in M}\!|W_i+\mathcal{F}(z_T)_i|  \!\right\}.
\end{align*}
It is noteworthy that two-sided detection will not affect the evaluation metric of the original non-watermarked latents $z_T$, as $\mathcal{F}(z_T)_i$ is a standard Gaussian distribution, which is symmetric with respect to sign changes. 
For watermarked latents, the matched pattern can still be extracted as we introduce the minimum operation. Therefore, two-sided detection is able to maintain detection accuracy when removal attacks are not engaged.

\begin{algorithm}[h]
\caption{Watermark Injection}
\label{alg:our_wm}
\begin{algorithmic}
\State{\bfseries Input:} 
Model $\mathcal{M}$, prompt $p$,  pattern $W$, parameter selector $f$, total generation step $T$.
\State {\bfseries Output:} 
Generated watermarked image $I^w$.
\State Sample noise $z_T$ from $\mathcal{N}(0,I)$.
\State $I^c\gets \mathcal{M}(p, z_T)$. \Comment{Generate a non-watermarked image}
\State $(t,l) \gets f(I^c)$
\Comment{Select time and location}
\State $z_t\gets \mathcal{M}_{T\to t}(p, z_T)$. \Comment{Denoise to step $t$}
\State $z_t^w\gets z_t\oplus \textup{Offset}(W,l)$ \Comment{Add at coordinate $l$}
\State $I^w\gets \mathcal{M}_{t\to 0}(p,z_t^w)$. \Comment{Denoise to image}
\end{algorithmic}
\end{algorithm}

\begin{algorithm}[h]
\caption{Watermark Detection}
\label{alg:our_detect}
\begin{algorithmic}
\State{\bfseries Input:} 
Suspicious image $I$, model $\mathcal{M}$, encoder $\mathcal{E}$, parameter selector $f$, watermarking pattern $W$, threshold $\tau$, watermarking mask $M$.

\State $z_0\gets \mathcal{E}(I)$, $(t,l)\gets f(I)$, $z_t\gets \mathcal{M}^{\textup{Inv}}_{0\to t}(z_0)$.
\State $W \gets \textup{Offset}(W, l)$ 
\State $d\gets \min \frac{1}{|M|} \left\{\sum\limits_{i\in M}|W_i-\mathcal{F}(z_{t,i})|,\sum\limits_{i\in M}|W_i+\mathcal{F}(z_{t,i})|  \right\}$

\Comment{Two-Sided Detection}
\If{$d < \tau$}
    \State \Return{Watermarked}
\Else
    \State \Return{Non-Watermarked}
\EndIf
\end{algorithmic}
\end{algorithm}

\section{Experiments}\label{sec:experiments}
\subsection{Experimental Setup}
\textbf{Datasets and models.}
In this work, we deploy Stable-Diffusion-2-1-base~\citep{rombach2022high} as the text-to-image diffusion model and use \citep{gustavo2024stable} as our text prompts to ensure a fair comparison with prior work~\citep{WenKGG23, ArabiWHC25}. Following the setting from \citet{MulleLTFQ24}, we select 100 pairs of watermarked and non-watermarked generated images for the evaluation of removal and forgery attacks. In non-adversarial scenarios, we measure on 1,000 pairs of images. 

\noindent \textbf{Baselines.}
We compare ISTS with recent content watermarking methods, including Tree-Ring~\citep{WenKGG23}, Shallow Diffuse~\citep{li2024shallow}, Gaussian-Shading~\citep{YangZCFZY24}, ROBIN~\citep{huang2024robin}, RingID~\citep{CiYSS25}, ZoDiac~\citep{ZhangLiBBG24}, and  SEAL~\citep{ArabiWHC25}. 
For removal and forgery attacks, we use imprinting attacks with gradient descent~\citep{MulleLTFQ24} (Imp-Removal/Forgery), simple averaging attacks~\citep{YangCSS24} (Avg-Removal/Forgery), and VAE attacks with gradient descent~\citep{JainKMTSMCMT25a} (VAE-Removal/Forgery).
Details of these attacks are provided in Appendix \ref{app:exp_detail}.

\noindent \textbf{Metrics.}
We deploy the area under the curve (AUC) of the receiver operating characteristic (ROC) curve, and the True Positive Rate (TPR) when the False Positive Rate (FPR) is 1\% (TPR@1\%FPR) as our detection metrics. 
For removal attacks, the detection metrics are evaluated between benign non-watermarked images and watermarked images after removal attacks. For forgery attacks, the detection metrics are evaluated between benign non-watermarked images and non-watermarked images after forgery attacks. Therefore, watermarking is more robust against removal attacks if the AUC and TPR@1\%FPR are higher, and is more robust against forgery attacks if the AUC and TPR@1\%FPR are lower. In practice, the detection threshold is fixed after the watermarking paradigm is proposed; thus, we utilize benign images as the non-watermarked criterion to maintain a consistent detection threshold for the TPR@1\%FPR metric.

\begin{table*}[t]
\small
\centering
\caption{The detection metric (AUC/TPR@1\%FPR) of various watermarking methods against removal attacks. ``Average" means the average AUC/TPR@1\%FPR across three removal attacks, Imp-Removal, Avg-Removal and VAE-Removal. ``Worst-Case" means the worst performance against these removal attacks, i.e., the lowest AUC and TPR@1\%FPR under a certain attack. Our ISTS demonstrates the strongest robustness against Imp-Removal and Avg-Removal, and outperforms all existing baselines in both averaged and worst-case scenarios for removal attacks.}
\label{tab:removal}
\begin{tabular}{lllllllllllll}
\toprule
Watermarking Method & \multicolumn{1}{c}{Original} & Imp-Removal  & Avg-Removal  & VAE-Removal& Average& Worst-Case\\
\midrule
Tree-Ring & 0.9999/1.00 & 0.2672/0.00& 0.5266/0.08& 0.9728/0.34  & 0.5889/0.14& 0.2672/0.00\\
Shallow Diffuse     & 1.0000/1.00 & 0.6752/0.00& 0.9730/0.49& 0.9987/0.93 & 0.8823/0.47& 0.6752/0.00\\
Gaussian Shading    & 1.0000/1.00 & 0.0000/0.00& 0.3707/0.01& \textbf{1.0000/1.00}  & 0.4569/0.34& 0.0000/0.00\\
ROBIN& 1.0000/1.00 & 0.0815/0.00& 0.7415/0.00& 0.9610/0.58  & 0.5947/0.19& 0.0815/0.00\\
RingID & 1.0000/1.00 & 0.0121/0.01& 0.4035/0.16& \textbf{1.0000/1.00}         &    0.4719/0.39      & 0.0121/0.01\\
Zodiac & 0.9999/1.00 & 0.0625/0.00& 0.2558/0.00& 0.7901/0.04  &   0.3695/0.01       & 0.0625/0.00\\
SEAL & 0.9998/1.00 &       0.5078/0.00& 0.9590/0.65 & 0.7884/0.37 &       0.7517/0.34&       0.5078/0.00\\
ISTS(Ours) & 1.0000/1.00 & \textbf{0.8210/0.18} & \textbf{0.9900/0.70} & 0.9979/0.85 & \textbf{0.9363/0.58} & \textbf{0.8210/0.18} \\
\bottomrule
\end{tabular}
\end{table*}

\begin{table*}[t]
\small
\centering
\caption{The detection metric (AUC/TPR@1\%FPR) of various watermarking methods against forgery attacks. 
``Average"  and ``Worst-Case" mean the average and the lowest AUC/TPR@1\%FPR across three forgery attacks, Imp-Forgery, Avg-Forgery and VAE-Forgery, respectively.
Our ISTS demonstrates the strongest robustness against Imp-Forgery, and outperforms all existing baselines in both averaged and worst-case scenarios for forgery attacks.}
\label{tab:forgery}
\begin{tabular}{lllllll}
\toprule
Watermarking Method & \multicolumn{1}{c}{Original} & Imp-Forgery   & Avg-Forgery  & VAE-Forgery        & Average & Worst-Case \\
\midrule
Tree-Ring & 0.9999/1.00 & 0.9914/0.73& 0.7932/0.01& 0.9525/0.48& 0.9124/0.41& 0.9914/0.73\\
Shallow Diffuse     & 1.0000/1.00 & 0.7621/0.11& 0.6285/0.02& 0.9533/0.55& 0.7813/0.23& 0.9533/0.55\\
Gaussian Shading    & 1.0000/1.00 & 1.0000/1.00& 0.4702/0.00& 1.0000/1.00& 0.8234/0.67& 1.0000/1.00\\
ROBIN& 1.0000/1.00 & 0.9995/0.95& 0.8784/0.03& 0.9290/0.29 & 0.9356/0.42& 0.9995/0.95\\
RingID & 1.0000/1.00 & 1.0000/1.00& 1.0000/1.00& 1.0000/1.00& 1.0000/1.00& 1.0000/1.00\\
Zodiac & 0.9999/1.00 & 1.0000/1.00& \textbf{0.2436/0.00} & 0.9541/0.42& 0.7326/0.47& 1.0000/1.00\\
SEAL & 0.9998/1.00 & 0.9536/0.48      &        0.4331/0.00 &\textbf{0.7213/0.04}&    0.7027/0.17 & 0.9536/0.48 \\
ISTS(Ours) & 1.0000/1.00 & \textbf{0.6340/0.00} & 0.4737/0.00& 0.9491/0.37& \textbf{0.6856/0.12} & \textbf{0.9491/0.37} \\
\bottomrule
\end{tabular}
\end{table*}

\subsection{Main Results}
\textbf{Results against removal attacks.}
We evaluate the detection AUC and TPR@1\%FPR with existing baseline watermarking, including Tree-Ring, Shallow Diffuse, Gaussian-Shading, ROBIN, RingID, Zodiac, SEAL, and our proposed ISTS, against three existing removal attacks against T2I diffusion watermarking: Imp-Removal, Avg-Removal, and VAE-Removal. 
The detailed results are provided in \Cref{tab:removal}. All these watermarking methods show strong detection performance when no removal attacks are involved (Original). However, after removal attacks, different watermarking methods demonstrate different utilities, where our ISTS achieves superior performance. 

In detail, for the Imp-Removal attack, almost all existing watermarking methods are severely degraded; the strongest baseline, Shallow Diffuse, only obtains an AUC of less than 0.7. Our watermarking improves the robustness with a significant gap, achieving over a 20\% AUC improvement (from 0.675 to 0.821) and an 18-fold TPR@1\%FPR enhancement (from 0.01 to 0.18). 
For Avg-Removal, Shallow Diffuse, SEAL, and our ISTS show strong robustness, where our method achieves the best AUC and TPR@1\%FPR. VAE-Removal is a relatively weak removal attack against which many existing watermarking techniques are robust, including Gaussian Shading and RingID, which are particularly effective. ISTS also demonstrates comparable robustness in the VAE-Removal attack.

We also make a general evaluation of these representative removal attacks with an average scenario and a worst-case scenario. In an average scenario, we assume that attackers do not have prior knowledge of these removal attacks against watermarking and take a random selection of the attack method against a certain watermarking. In this case, we evaluate the detection performance by calculating the average AUC and TPR@1\%FPR. Our ISTS achieves the best robustness in this scenario, with over 0.93 AUC and 0.58 TPR@1\%FPR. In the worst-case scenario, we assume that attackers can choose the best removal method against certain watermarking, and the detection metrics are evaluated by finding the lowest AUC and TPR@1\%FPR across these attacks. In this scenario, we find that Imp-Removal dominates, and the superior performance of ISTS against Imp-Removal induces the best robustness in the worst-case scenario.

\noindent \textbf{Results against forgery attacks.}
Similar to removal attacks, we also evaluate the detection AUC and TPR@1\%FPR for various content watermarking methods against three existing forgery attacks, Imp-Forgery, Avg-Forgery, and VAE-Forgery. The detailed results are provided in  \Cref{tab:forgery}. It is noteworthy that, unlike removal attacks, lower AUC/TPR@1\%FPR indicates better watermarking against forgery attacks, as the watermarked images are harder to be forged. For Imp-Forgery, all existing methods except Shallow Diffuse are easily forged (over 0.95 AUC), and ISTS obtains superior performance, achieving over 20\% improvement on AUC compared with the strongest baseline, Shallow Diffuse (0.76 to 0.63). Avg-Forgery is a relatively weak attack, with almost all existing watermarking (except RingID) having good resistance against it. Zodiac performs best, and ISTS also gets comparable results. VAE-Forgery seems stronger against all existing methods, SEAL dominates this attack, ISTS achieves comparable results with some baselines like Tree-Ring, Shallow Diffuse, ROBIN and Zodiac. 

In the average and worst-case scenarios, our ISTS still demonstrate superiority across three forgery attacks, with the lowest AUC (0.685 and 0.949 for average and worst-case scenarios) and TPR@1\%FPR (0.12 and 0.37 for average and worst-case scenarios).

It is noteworthy that, although SEAL outperforms against VAE-Forgery and achieves comparable results with our ISTS in the average and worst-case scenarios, their poor image quality and robustness against image distortions and purifications will limit their applicability, as demonstrated in the next section.


\subsection{More Experimental Results}



\begin{figure}[h]
    \centering
    \includegraphics[width=0.99\linewidth]{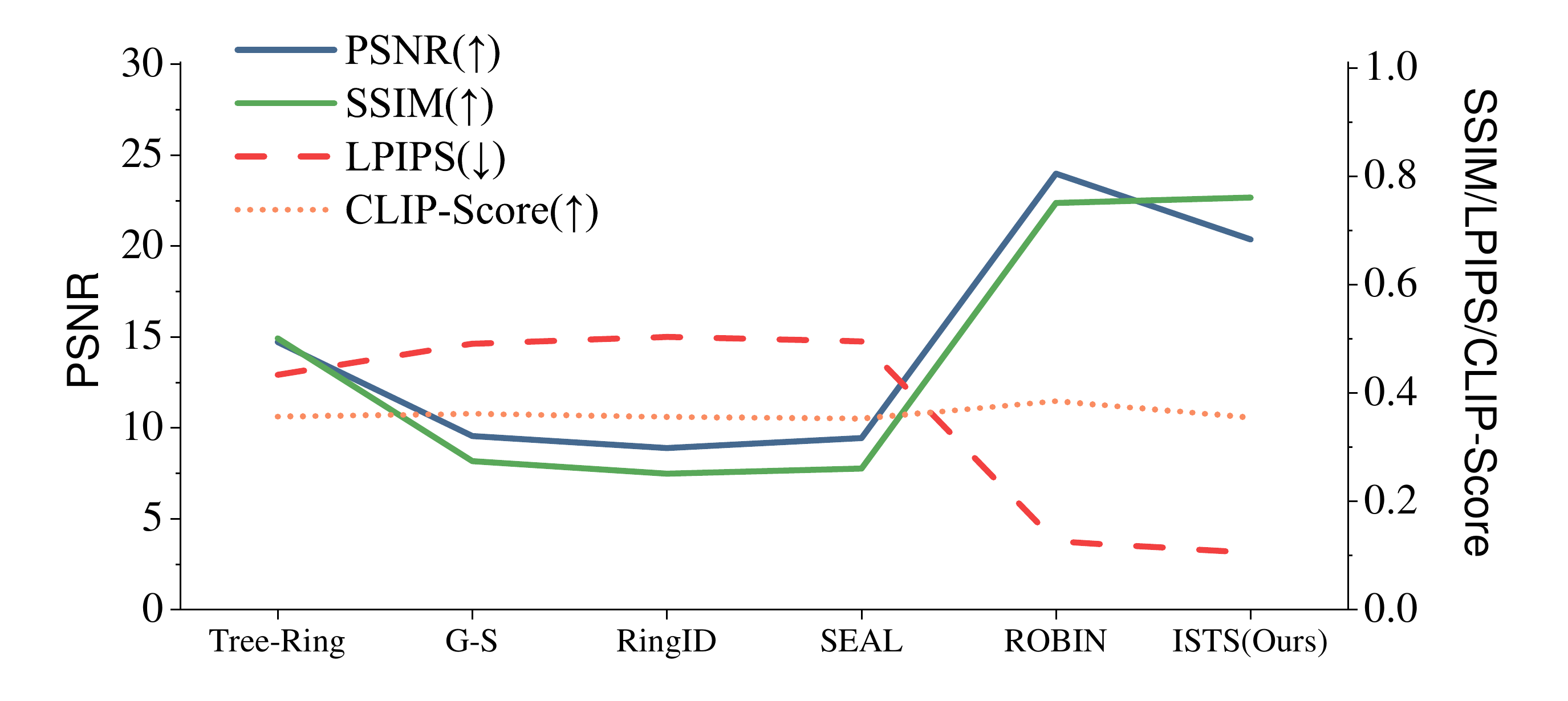}
    \caption{The PSNR, SSIM, LPIPS and CLIP-Score for various content watermarking. Our ISTS achieves comparable results with ROBIN and outperforms other methods. G-S means Guassian-Shading.}
    \label{fig:img_qual}
\end{figure}

\noindent \textbf{Image quality and semantics.}
We evaluate the generated image quality with metrics including Peak Signal-to-Noise Ratio (PSNR), Structural Similarity Index (SSIM)~\citep{wang2004image}, and Learned Perceptual Image Patch Similarity (LPIPS)~\citep{zhang2018unreasonable}. Furthermore, we also measure the CLIP-Score~\citep{radford2021learning} between generated images and prompts using OpenCLIP-ViT/G~\citep{cherti2023reproducible}. 
Results shown in \Cref{fig:img_qual} demonstrate that for PSNR, SSIM, and LPIPS, our ISTS outperforms various content watermarking methods like Tree-ring, Gaussian-Shading, RingID, and SEAL, and achieves comparable results with ROBIN. For CLIP-Score, existing methods have similar results, implying good semantic alignment with text prompts.
It is worth noting that, although ROBIN has comparable image quality with ISTS, it is very fragile to removal and forgery attacks as displayed in \Cref{tab:removal,tab:forgery}.

\begin{figure}[h]
    \centering
    \includegraphics[width=0.99\linewidth]{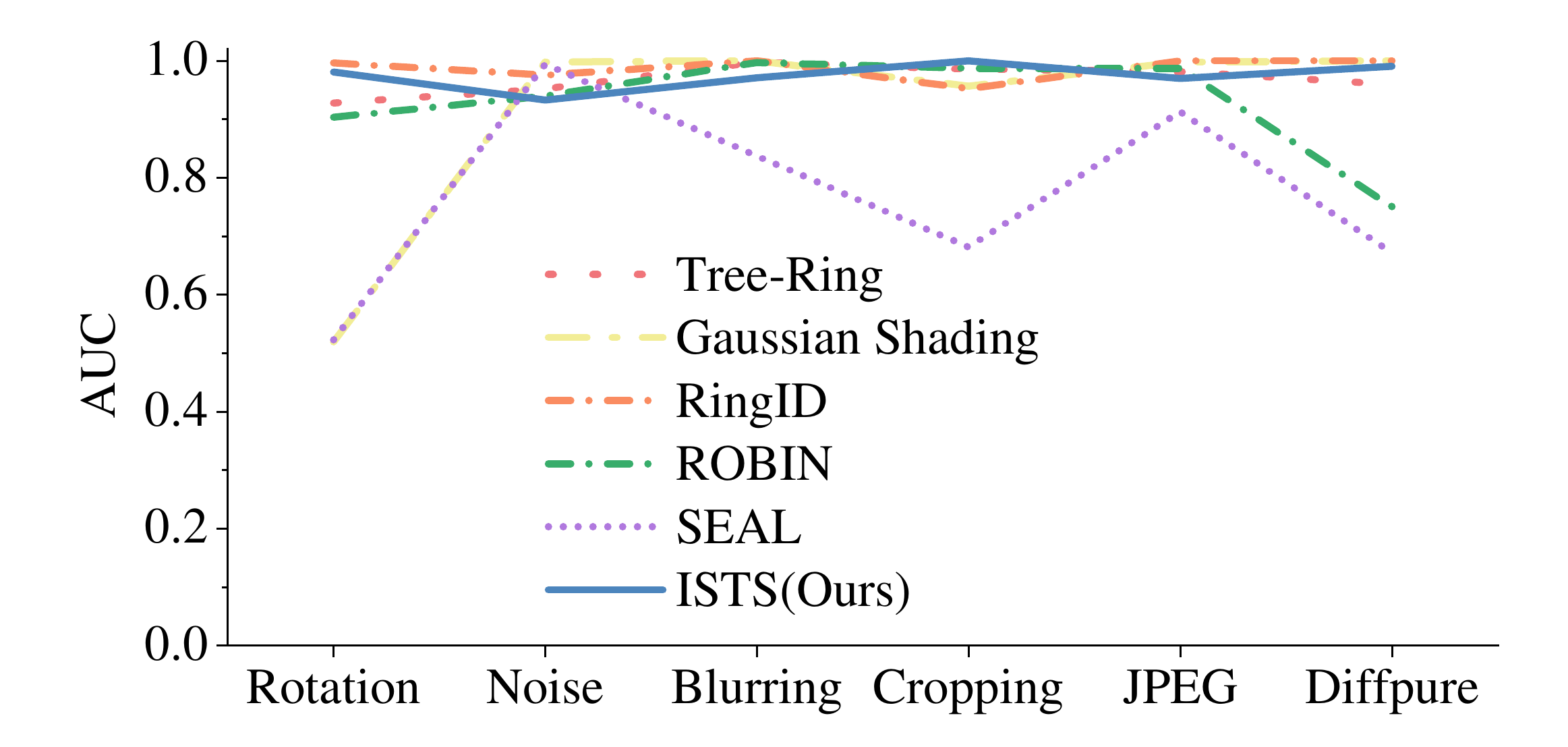}
    \caption{Watermarking detection AUC under various image distortions (Rotation, Noise, Blurring, Cropping, JPEG) and diffusion purification (Diffpure). Results demonstrate that our method achieves high AUC across these data augmentations.}
    \label{fig:img_distort}
\end{figure}

\noindent \textbf{Robustness against image distortions and purifications.}
To further investigate the robustness of content watermarking, followed the setting of \citet{WenKGG23}, we evaluate them on several well-known image distortions, including $75^{\circ}$ rotation (Rotation), $\sigma=0.1$ Gaussian noise (Noise), Gaussian blur with $8\times8$ filter size (Blurring), $75\%$ random cropping and scaling (Cropping), and $25\%$ JPEG compression (JPEG). Moreover, we also evaluate the effectiveness of watermarking under reconstruction through diffusion purification~\citep{zhao2024invisible} (Diffpure). Results in \Cref{fig:img_distort} show that Tree-Ring, RingID ,and our ISTS obtain decent robustness against these image distortions. Gaussian-Shading is vulnerable to Rotation, ROBIN is fragile to Diffpure, and SEAL becomes ineffective under Rotation, Blurring, Cropping, and Diffpure.

To further quantify the robustness for watermarking, we provide the detection AUC in both average and worst-case scenarios across these distortions on \Cref{tab:distort}. ISTS possesses comparable robustness with Tree-Ring and RingID, and outperforms Gaussian Shading, ROBIN, and SEAL. It is noteworthy that although SEAL is comparable to our watermarking against forgery attacks, the poor robustness against image distortions restricts its real-world applications.
Furthermore, even though RingID gains superior performance under image distortions, the weakness against removal and forgery attacks, as demonstrated in \Cref{tab:removal,tab:forgery} restricts their real-world applications.

\begin{table}[h]
\caption{The average and worst-case detection AUC across Rotation, Noise, Blurring, Cropping, JPEG, and Diffpure. T-R means Tree-Ring, G-S means Gaussian-Shading.}
\small
\label{tab:distort}
\hspace{-4pt}
\setlength{\tabcolsep}{2pt}
\begin{tabular}{lllllll}
\toprule
AUC      & T-R & RingID & G-S & ROBIN  & SEAL   & ISTS(Ours)   \\
\midrule
Average    & 0.9750     & 0.9874 & 0.9120& 0.9276 & 0.7699 & 0.9742 \\
Worst-Case & 0.9276    & 0.9526 & 0.5194 & 0.7504 & 0.5228 & 0.9331 \\
\bottomrule
\end{tabular}
\end{table}

\noindent \textbf{Ablation study of each component.} To investigate the influence of each component in our ISTS, we conduct an ablation study in \Cref{tab:comp} to evaluate the performance without certain components. Results demonstrate that, for all removal and forgery attacks except Avg-Forgery, combining all three components achieves the best robustness. Notably, dynamic patterns are important for robustness against forgery attacks, especially for Imp-Forgery (0.72 to 0.62) and Avg-Forgery (0.60 to 0.47), because attackers can easily forge the fixed watermarking pattern even with a single image, as discussed in \Cref{sec:dyn_pattern}. Two-sided detection contributes greater robustness against Imp-Removal (0.71 to 0.82) as it resists attacks on opposite latent representations, which validates our findings.

\begin{table}[h]
\caption{Detection AUC of components in our ISTS watermarking. ``w/o Dyn-Pattern", ``w/o Dyn-Step" and ``w/o Two-Sided" represent removing the pattern position selection, removing the injection timestep selection, and removing the two-sided detection back to one-sided, respectively.}
\small
\label{tab:comp}
\hspace{-4pt}
\setlength{\tabcolsep}{2pt}
\begin{tabular}{l|lll|llll}
\toprule
\multirow{2}{*}{Method}  & \multicolumn{3}{c|}{Removal($\uparrow$)}      & \multicolumn{3}{c}{Forgery($\downarrow$)}      \\ \cline{2-7}
 & Imp  & Avg      & VAE       & Imp  & Avg      & VAE         \\
\midrule
ISTS & \textbf{0.8210} & \textbf{0.9900} & \textbf{0.9979} & \textbf{0.6340} & 0.4737        & \textbf{0.9491} \\
w/o Dyn-Pattern&     0.7893& 0.9762         & 0.9969         & 0.7202         & 0.6014         & 0.9625         \\
w/o Dyn-Injection & 0.8014& 0.9809         & 0.9973 &0.6678& \textbf{0.4440} & 0.9515      \\
w/o Two-Sided & 0.7129        & 0.9866         & 0.9945         &         0.6412& 0.4572         & 0.9532   \\
\bottomrule
\end{tabular}
\end{table}

\section{Conclusion}
In this paper, a novel content watermarking method against both removal and forgery attacks called ISTS has been proposed. ISTS utilizes instance-specific parameters to control the watermarking pattern and the injection timestep, and calculates the metric with two-sided detection. Experiments on various removal and forgery attacks have demonstrated the superior performance against existing methods.

\textbf{Limitations and future works.}
Although our ISTS manifests the state-of-the-art performance against removal and forgery attacks, the worst-case robustness is still not satisfactory for effective removal and forgery attacks such as Imp-Removal and VAE-Forgery. As a provenance and credit attribution approach, more robust content watermarking should be developed for real-world applications.

\textbf{Codes.}
Please find our codes and implementation details at https://github.com/hala64/ISTS.

\clearpage

\section*{Acknowledgement}
This paper is supported by the Strategic Priority Research Program of CAS Grant XDA0480502, the Robotic AI-Scientist Platform of the Chinese Academy of Sciences, and NSFC Grants 92270001 and 12288201.

    \small

\clearpage
\onecolumn
\appendix
\section{Appendix of Paper "Towards Robust Content Watermarking Against Removal and Forgery Attacks"}\label{sec:appendix}

\subsection{Experimental Details}
\label{app:exp_detail}

\noindent \textbf{Watermarking hyperparameter assignment.}
After gaining the label $y^c\in[C]$ for $C$-classes clustering, we need to assign corresponding injection timestep $t$ and pattern location offset $l$.
Specifically, we denote the injection timestep range in $[T_1, T_2]$, the pattern location offset $l=(l_x,l_y)$ range in $[l_x^1,l_x^2]\times[\l_y^1,l_y^2]$.
The injection timestep $t$ is set as $$t=T_1+\left[y^c {\rm \,\, mod} (T_2-T_1)\right],$$
the pattern position offset is set as 
$$l_x=l_x^1+\left[y^c {\rm \,\, mod} (l_x^2-l_x^1)\right], l_y=l_y^1+\left[ \left(y^c//(l_y^2-l_y^1)\right) {\rm \,\, mod} (l_y^2-l_y^1)\right].$$
The watermarking pattern after location offset should be 
$$\textup{Offset}(W,l)_{i,j}=W_{i+l_x,j+l_y}.$$

\noindent In this paper, we set the clustering class $C=1024$, the timestep range in $[10,20]$, and the pattern position offset be $[-12,12]^2$ by default.

\noindent \textbf{Watermark injection.} 
After getting the watermarking pattern $W^o=\textup{Offset}(W,l)$, we add the watermarking pattern into the $t$-step latent $z_t$. 
Building upon with the method proposed by \citet{WenKGG23}, denote the watermarking mask in the frequency domain be $M$, $M$ is also deployed with location offset $l$ similar as $W$, denoted as $$M^o_{i,j}=M_{i+l_x,j+l_y}.$$
The watermarking channel is 0 and the radius is set to be 20. Then we replace the corresponding pixel $k=(i_k,j_k)$ of $z_t$ by $W^o$ under mask $M^o$ in the frequency domain, i.e. 
\begin{align}
\mathcal{F}(z_t)_k =
\begin{cases}
W^o_k, & \text{if \,\,} k \in M^o,\\[4pt]
\mathcal{F}(\mathcal{M}_{T\to t}(p, z_T))_k, & \text{otherwise.}
\end{cases} \nonumber
\end{align}
Finally, the watermarked $t$-step latent $z_t^w$ is obtained by inverse Fourier transformation:
$$z_t^w=\mathcal{F}^{-1}(\mathcal{F}(z_t)),$$
after the modification on $\mathcal{F}(z_t)$.

\noindent To make the notation simple, in our Algorithm \ref{alg:our_wm}, we donate the above injection process as $$z_t^w=z_t\oplus \textup{Offset}(W,l).$$

\noindent \textbf{Parameter selector.} For a new prompt $p_{new}$ and the corresponding generated image $I_{new}$, we need to assign its watermarking parameter with our parameter selector $f=h\circ g$, where $g$ is a pre-trained CLIP feature extractor, and $f$ is the classifier trained by existing features and their assigned labels. 

\noindent To reduce the computational complexity, we use a very simple two-layer neural network to align extracted features with their watermarking parameters, where the hidden dimension is same as the input feature dimension, and apply ReLU activation, Batch Normalization with $p=0.5$ dropout rate.

\subsection{Details on Removal and Forgery Attacks}
\label{app:removal_forgery}
\textbf{Imp-Removal.}
In \citet{MulleLTFQ24}, they try to remove watermarks by optimizing the follow equation:
$$\mathcal{L}(\delta)=\|\mathcal{M}_{0\to T}(p, z_0^w+\delta)+z_T^w\|_2,$$
where $\mathcal{M}$ is the surrogate diffusion model, $p$ is the text prompt, $z_0^w=\mathcal{E}(x^w)$ derive from the obtained watermarked image $x^w$ with encoder $\mathcal{E}$, $z_T^w=\mathcal{M}_{0\to T}(p, z_0^w)$ is an estimation of $T$-step latent after DDIM inversion from $z_0^w$. Tne Imp-Removal attack aims to induce watermarked latent to the opposite position in timestep $T$, which could be vulnerable to our two-sided detection. Finally, the image after Imp-Removal attack $\hat{x}^w$ is decoded by decoder $\mathcal{D}$ from the perturbed latent:
$$\hat{x}^w=\mathcal{D}(z_0^w+\delta).$$

\noindent Following the setting provided in \citet{MulleLTFQ24}, we set the optimization steps be 150, with learning rate be 0.01. Furthermore, it is noteworthy that, \citet{MulleLTFQ24} considers the surrogate model scenario, where the attack model be Stable-Diffusion 2.1 (SD 2.1), and the victim models be different, like SD 2.1-Anime (SD 2.1 Finetuned on Anime), Stable Diffusion XL~\citep{podell2023sdxl}, etc. In this paper, we consider a stronger case for attackers, that they can use the same diffusion model as surrogate, inducing a white-box rather than grey-box scenario. As our evaluation based on this white-box scenario, the robustness of our watermarking will be not bad than those from grey-box scenario.

\noindent \textbf{Imp-Forgery.}
In \citet{MulleLTFQ24}, they want to forge existing watermarking method by optimizing the following loss:
$$\mathcal{L}(\delta)=\|\mathcal{M}_{0\to T}(p, z_0^c+\delta)-z_T^w\|_2,$$
where $z_0^c=\mathcal{E}(x^c)$ derive from the clean image $x^c$ ready to be forged, $z_T^w=\mathcal{M}_{0\to T}(p, z_0^w)$ is an estimation of $T$-step latent after DDIM inversion from $z_0^w$, and $z_0^w=\mathcal{E}(x^w)$ derive from the obtained watermarked image $x^w$. Other notations are similar to Imp-Removal. Similarly, the optimization steps are 150 with learning rate is 0.01 following the setting of \citet{MulleLTFQ24}. We also evaluate on the white-box scenario in Imp-Forgery attack. The image after Imp-Forgery attack $\hat{x}^w$ is decoded by decoder $\mathcal{D}$ from the perturbed latent:
$$\hat{x}^w=\mathcal{D}(z_0^c+\delta).$$

\noindent \textbf{Avg-Removal and Avg-Forgery.}
In \citet{YangCSS24}, they find that watermarking patterns can be revealed by averaging a collection of watermarked and non-watermarked images. Specifically, they propose averaging over $N$ images and calculating their residual as:
$$\delta=\frac{1}{N} \left(\sum\limits_{i=1}^N x_{w,i}-\sum\limits_{i=1}^N x_{c,i}\right),$$
where $\{x_{c,i}\}_{i\in[N]}$ are clean non-watermarked images, $\{x_{w,i}\}_{i\in[N]}$ are watermarked images.

\noindent In Avg-Removal attack, they modify the watermarked image $x^w$ to $\hat{x}^w$ by
$$\hat{x}^w=x^w-\delta,$$
where in Avg-Forgery attacks, they modify the clean image $x^c$ to $\hat{x}^w$ by
$$\hat{x}^w=x^c+\delta.$$
In this paper, we first generate 100 pairs of non-watermarked and watermarked images to extract the residual $\delta$, then conduct Avg-Removal and Avg-Forgery attacks for each watermarked/non-watermarked image.

\noindent \textbf{VAE-Removal.}
In \citet{JainKMTSMCMT25a}, similar to \citet{MulleLTFQ24}, they try to remove and forge watermarks with a single watermarked image. Unlike \citet{MulleLTFQ24} uses a surrogate diffusion model to optimize perturbations in the $T$-step noise latent space, \citet{JainKMTSMCMT25a} directly optimize attacks on latent image space with only the surrogate VAE encoder. Specifically, in VAE-Removal attack, they optimize injected noise $\delta$ by:
$$\min\limits_{\delta}\|\mathcal{E}(x^w+\delta)-\mathcal{E}(\mu_{x^w})\|_2 +\lambda\|\delta\|_2,$$
where $x^w$ is the watermarked image ready to be removed, $\mathcal{E}$ is the VAE encoder, $\lambda$ is the balancing hyperparameter, and $\mu_{x^w}$ is the plain image with all values equal to the mean of the watermarked image $x^w$.

\noindent In this paper, followed the setting of \citet{JainKMTSMCMT25a}, we choose the VAE encoder from Stable Diffusion v1.4, and set $\lambda$ be $5\times 10^{4}$.

\noindent \textbf{VAE-Forgery.}
Similar to VAE-Removal attacks, VAE-Forgery proposed by \citet{JainKMTSMCMT25a} also optimize the perturbation at the latent image space. Specifically, VAE-Forgery works as:
$$\min\limits_{\delta}\|\mathcal{E}(x^c+\delta)-\mathcal{E}(x^w)\|_2 +\lambda\|\delta\|_2,$$
where $x^c$ is the clean non-watermarked image ready to be forged, $x^w$ is the obtained watermarked image, other notations are similar to VAE-Removal. 

\noindent Same as VAE-Removal, the VAE encoder is from Stable Diffusion v1.4, and the hyperparameter $\lambda=5\times10^4$.


\subsection{Different Intermediate Steps for Watermarking Injection}
\begin{figure}[h]
    \centering
    \includegraphics[width=0.5\linewidth]{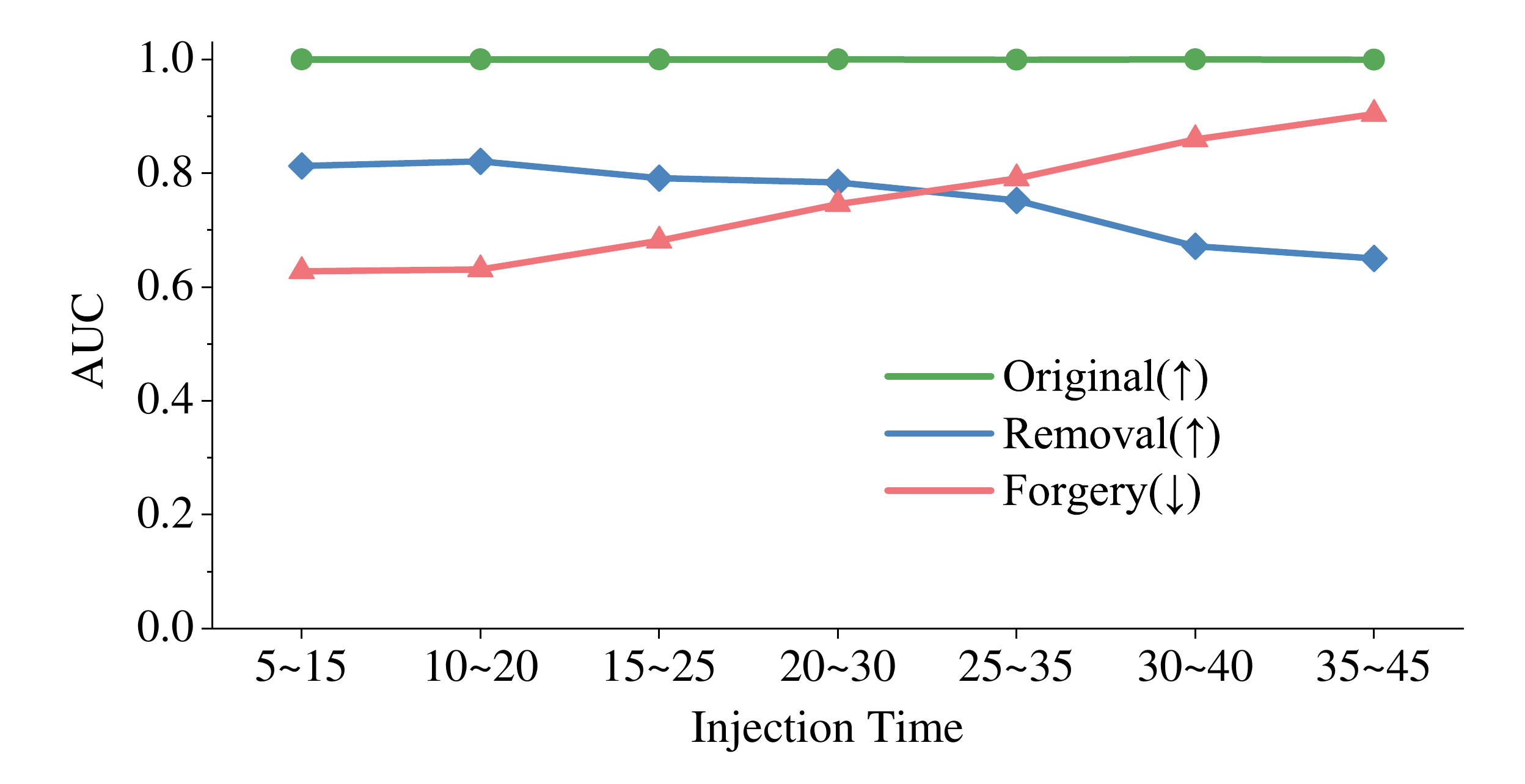}
    \caption{Different injection steps.}
    \label{fig:steps}
\end{figure}

To further investigate the impact of intermediate steps for watermarking injection, we evaluate different ranges of injection steps from $t=5$ to $t=45$, the overall diffusion steps is $T=50$. As our watermarking injection dynamically select $t\in[10,20]$, to ensure the consistency, we evaluate different steps with the same range of variation, from $[5,15]$ to $[35,45]$. We use Imp-Removal and Imp-Forgery as the removal and forgery attacks, results are shown in Figure \ref{fig:steps}. 

\noindent It reveals that although the detection AUC remains sufficiently high for original watermarking without attacks across different intermediate steps, the performance under removal and forgery attacks have changed considerably. As steps go larger, Detection AUC becomes worse for both removal and forgery attacks, implying a relative small injection steps when designing robust watermarking.

\subsection{Potential Side-Channel Attacks on ISTS}
A simple side-channel attack of our ISTS watermarking is the potential leakage of the public CLIP extractors and even the parameter selector model.
However, we believe that information extracted from CLIP features or the parameter selector model of potentially leaked watermarked images does not compromise the security of our ISTS watermarking scheme. 
This is because the assigned labels $y_p^c$ can be obfuscated using a (pseudo-)random permutation $\mathcal{R}$ controlled by a secret key. Given a large number of clusters (e.g., 1024), it becomes infeasible for an attacker to recover the injection parameters $(t, l)$, as they cannot infer the permuted label $\mathcal{R}(y_p^c)$ because $2^{1024}$ is a infeasible large number for any attackers. 
Moreover, in practical deployments, watermarking methods are typically applied to proprietary models, where adversaries are limited to black-box access via APIs and do not have visibility into the model’s internal structures or algorithms.



\end{document}